# Semantic Similarity Measures Applied to an Ontology for Human-Like Interaction

**Esperanza Albacete**                                      EALBACET@INF.UC3M.ES
**Javier Calle**                                                FCALLE@INF.UC3M.ES
**Elena Castro**                                             ECASTRO@INF.UC3M.ES
**Dolores Cuadra**                                          DCUADRA@INF.UC3M.ES
*Computer Science Department, Carlos III University,*
*Madrid 28911, Spain*

## Abstract

The focus of this paper is the calculation of similarity between two concepts from an ontology for a Human-Like Interaction system. In order to facilitate this calculation, a similarity function is proposed based on five dimensions (sort, compositional, essential, restrictive and descriptive) constituting the structure of ontological knowledge. The paper includes a proposal for computing a similarity function for each dimension of knowledge. Later on, the similarity values obtained are weighted and aggregated to obtain a global similarity measure. In order to calculate those weights associated to each dimension, four training methods have been proposed. The training methods differ in the element to fit: the user, concepts or pairs of concepts, and a hybrid approach. For evaluating the proposal, the knowledge base was fed from WordNet and extended by using a knowledge editing toolkit (Cognos). The evaluation of the proposal is carried out through the comparison of system responses with those given by human test subjects, both providing a measure of the soundness of the procedure and revealing ways in which the proposal may be improved.

## 1. Introduction

The main purpose of an ontology in a human-like interaction system is to unify the representation of each concept, relating it to the appropriate terms, as well as to other concepts with which it shares a semantic relation. Furthermore, the ontological component should also be able to perform certain inferential processes, such as the calculation of semantic similarity between concepts. The subject of similarity has been and continues to be widely studied in the fields and literature of computer science, artificial intelligence, psychology and linguistics. Good similarity measures are necessary for several techniques from these fields including information retrieval, clustering, data-mining, sense disambiguation, ontology translation and automatic schema matching. The present paper focuses on the study of semantic similarity between concepts in an ontology from the framework of natural interaction.

The principal benefit gained from this procedure is the ability to substitute one concept for another based on a calculation of the similarity of the two, given specific circumstances. From the user's perspective, the procedure allows for the use of synonyms (terms related to a single concept) of a concept in the case where the user is not familiar with the original concept itself. Moreover, semantic similarity offers the possibility to build explanations for clarifying a concept to the user based on similar concepts, thereby enhancing communicative effectiveness.

On the other hand, the system may also be able to understand a previously-unknown concept, as long as the user is able to relate it to similar concepts that are previously known by the system. In this way, the system will learn new concepts and automatically enrich its ontology to improve future interactions.

The first task of this study is to develop a semantic similarity measure that takes into account particular ontological dimensions described in an earlier study (Calle, Castro & Cuadra, 2008). In this approach, the conceptualization comprises seven ontological dimensions: semiotic, sort, compositional, essential, restrictive, descriptive, and comparative. The first three dimensions have been previously applied in related works, as will be stated in Section 2. Essential, restrictive and descriptive dimensions are part of the nature of the concept, can influence human judgment of similarity and will be detailed in Section 3. The seventh one, comparative dimension, is derived from previous dimensions and is in charge of calculating the degree of similarity between ontological concepts.

The second goal of the present article is to evaluate the quality of the mechanism developed for the calculation of similarities between two concepts in an ontology which is specially designed for a human-like interaction system (Calle F., 2004). To achieve this, several experiments have been designed and performed here. Before these experiments and the consequent evaluation of the semantic similarity measure can be carried out, however, it is necessary to implement the similarity dimensions defined in the conceptual model and feed the database with a large number of concepts.

To briefly outline the content that follows in this paper, Section 2 reviews the literature on similarity measures in ontologies and the methods available for their evaluation. In Section 3, an approach to similarity measures applied to an ontological model based on several dimensions is proposed. In Section 4, a detailed explanation is provided of the experiments designed to test the proposal, as well as the results obtained from their execution. Section 5 discusses the limitations encountered in the study. Finally, Section 6 presents conclusions for future research.

## 2. Related Work

The present section of this paper has two main objectives. First, it aims to provide an overview of the different types of approaches available for the comparison of concepts in ontologies and, in so doing, to identify the foundations on which the desired similarity measure may be modeled, taking into account the seven dimensions described in a previous study (Calle et al., 2008). Secondly, it aims to select the best way to evaluate the results yielded from this desired similarity measure according to other studies regarding similarity metrics assessment.

Basically two types of methods exist for the comparison of terms in a graph-based ontology: edge-based methods using graph edges and their types as the data source and node-based methods using graph nodes and their properties as the main data source. The simplest and most intuitive similarity measure, the former method is based mainly on the counting of the number of edges in a path between two terms on a graph (Rada, Mili, Bicknell & Blettner, 1989). Within the edge-based method, two general approaches exist: firstly, a distance approach that selects either the shortest path or the average of all paths (when more than one path exists) and secondly a common path approach that calculates similarity directly by the length of the path from the lowest common ancestor of the two terms to the root node (Wu & Palmer, 1994). Over the past few years, a variety of edge-based methods have been defined (Resnik, 1995; Leacock & Chodorow, 1998).

All edge-based methods are grounded in two basic assumptions: firstly, that nodes and links are uniformly distributed in the ontology, that is, terms at the same depth have the same specificity (Budanitsky, 1999) and, secondly, that edges at the same level in the ontology indicate the same semantic distance between terms. However, these suppositions are rarely true in the majority of ontologies. For this reason, several strategies have been proposed in response to this fact. One example of such a strategy is the weighting of edges according to their hierarchical depth or the use of node density and link type (Richardson, Smeaton & Murphy, 1994). Nevertheless, these strategies do not solve the aforementioned problems due to the fact that terms at the same depth do not necessarily have the same specificity and that edges at the same level do not necessarily represent the same semantic distance.

The second, or node-based, method relies on the comparison of the properties of the terms involved which can be related to the terms themselves, their ancestors or their descendants. A commonly used concept in these methods is that of information content (IC), providing a measure of how specific and informative a term is. The IC of a term c can be quantified as the negative log-likelihood, IC = -log p(c), where p(c) is the probability of the occurrence of c in a specific corpus, generally being estimated by its annotation frequency. Another approach employed to obtain the IC is based on the number of children a term has in the ontological structure (Seco, Veale & Hayes, 2004). The concept of IC can be applied to the common ancestors of two terms in order to quantify the information they share and, thereby, measure their semantic similarity. In this way, two main approaches exist. The first is the most informative common ancestor (MICA) technique in which only the common ancestor with the highest IC is considered (Resnik, 1995). The second is the disjoint common ancestor (DCA) technique in which all disjoint common ancestors are considered (the common ancestors that do not subsume any other common ancestor). In one definition (Lin, 1998), the similarity between two concepts using the node-based method has been expressed as the ratio between the amount of information needed to state the commonality between the two concepts and the information needed to fully describe them. Moreover, a similarity measure for hierarchical ontologies called ontology structure-based similarity (OSS) has also been defined (Schickel-Zuber, 2007) and whose major ingredient is the computation of an a-priori score of a concept c, (APS(c)), which shares some similarities with IC (i.e., both are calculated from the topology and structure of the ontology reflecting the information contained within and between the concepts).

Additionally, several hybrid methods have also been defined in an attempt to improve the results of both techniques defined above. In the work of Jiang and Conrath, (1997), for example, a combined model is defined that is derived from the edge-based notion by adding information content as a decision factor. The link strength between two concepts is defined as the difference of information content between them.

With the aim of collecting all different methods and approaches, SimPack, a generic Java library of similarity measures for use in ontologies, has been created (Bernstein, Kaufmann,

Kiefer & Bürki, 2005) and includes the implementation of ontology-based similarity methods (including edge-based and node-based measures). It is important to note that the majority of the techniques described to define semantic similarity between concepts have been applied to hierarchical ontologies whose structure takes into account only one or two dimensions in the same graph. For example, WordNet (Fellbaum, 1998) consists of an ontological graph with over 100,000 concepts and whose edges model the is_a and part_of relationships. A Perl module (Pedersen, Patwardhan & Michelizzi, 2004) was implemented for this lexical database with a variety of semantic similarity measures. Another example of application is the Gene Ontology (Department of Genetics, Stanford University School of Medicine, California, USA., 2000), one of the most important ontologies within the bioinformatics community, with over 20,000 concepts and modeling is_a and part_of relationships in the same graph. Thus, while none of the techniques described in this section can be supposed to be appropriate in dealing with more than two dimensions of similarity, they can nevertheless be useful to attempt to define some of the dimensions in the present study's ontological model.

The second aim of the present section is to review the assessment techniques for ontological similarity functions used in earlier studies. The gold standard established in the majority of the experimental evaluations of similarity (Resnik, 1999; Jiang & Conrath, 1997; Altintas, Karsligil, & Coskun, 2005; Schickel-Zuber, 2007; Bernstein et al., 2005) is based on the experiment described in Miller and Charles' study (1991) which has become the benchmark for determining the similarity of words in natural language processing research. This experiment relies on the similarity assessments made by 38 university students when provided with 30 name pairs chosen a priori to cover high, intermediate and low levels of similarity and when asked to assess the similarity of their meaning on a scale from 0 (no similarity) to 4 (perfect synonymy). The average of scored values represents a good estimation of the degree of similarity between two terms.

In certain evaluations based on human judgment (Inkpen, 2007; Bernstein et al., 2005), variations in the number of participants or the way to administer the questionnaire have been introduced. In one of these studies (Bernstein et al., 2005), a website containing a survey tool was designed to perform the evaluation. In the Web experiment, subjects were asked to assess the similarity between 73 pairs of concepts on a scale from 1 (no similarity) to 5 (identical). Finally, subjects were also given the possibility of adding comments to their assessment. To evaluate the quality of the similarity measures, its results were compared with the test subjects' assessments using the corrected Spearman rank correlation coefficient.

It can be concluded that human reasoning is one of the most widely-used methods of comparison when performing validation of a similarity measure. For this reason, such a methodology has also been used in the experimentation section of the present study. Since it is difficult to run a user-based evaluation with complicated ontologies, for example, the Gene Ontology (Lord, Stevens, Brass & Goble, 2003), it has been deemed necessary here to find or model an ontology with elements that test subjects could understand. Therefore, once the ontological module is implemented, it must be populated with a sufficiently good coverage of domain knowledge, that is, enough knowledge to meet the system requirements.

## 3. Theoretical Approach

The conceptual model grounding the present study (Calle et al., 2008) distributes ontological knowledge into seven different dimensions. The semiotic dimension represents the relationship between concepts, terms and language. For example, as shown in Figure 1, the concept of the WordNet's synset 3082979 corresponds to a machine that is able to perform calculations automatically, and one of the terms associated with this concept is "computer". Other terms related to this concept are "computing machine", "computing device", "data processor", "electronic computer" and "information processing system", all of them also linked to the concept that corresponds to the English language (synset 6947032).

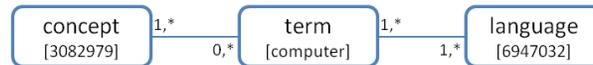

**Figure 1: Example of semiotic dimension representation**

The *sort dimension* represents the is_a relationship between concepts, relates each concept with other concepts and models a polytree structure. For instance, as shown in Figure 2, the terms "node", "server" and "web site" are related to concepts that are instances of "computer".

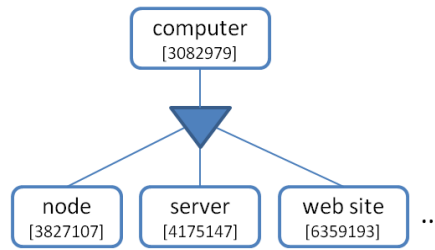

**Figure 2: Sort dimension example**

The *essential dimension* represents the general taxonomy of concepts. This taxonomy is located in the nodes at the top of the polytree represented in the sort dimension. Therefore, the relations included in its design are already observed in the sort dimension. But since they organize the knowledge at the higher abstraction level (they are more discriminative) they should be taken into account separately, adding extra value to similarity measure.

Its design is crucial for attaining good similarity measures, and determines the usefulness of this dimension. The essential dimension of WordNet (Princeton Univ., 2011), for example, classifies the concepts into four main linguistic categories (verb, noun, adjective, adverb). Such approach is the most adequate for a linguistic interaction domain, but may be weaker in a general interaction domain. This proposal includes an essential design inspired in previous (Calle et al., 2008) and related works (Gee, 1999; Miller, 1995) and refined through preliminary experimentation. The design departs from three main categories (abstract, actions and entities) and develops main classes of concepts, as shown in Figure 3. Finally, it should be added that this proposal is aimed to general interaction domains, and could be improved if suited to specific domains for particular interaction systems.

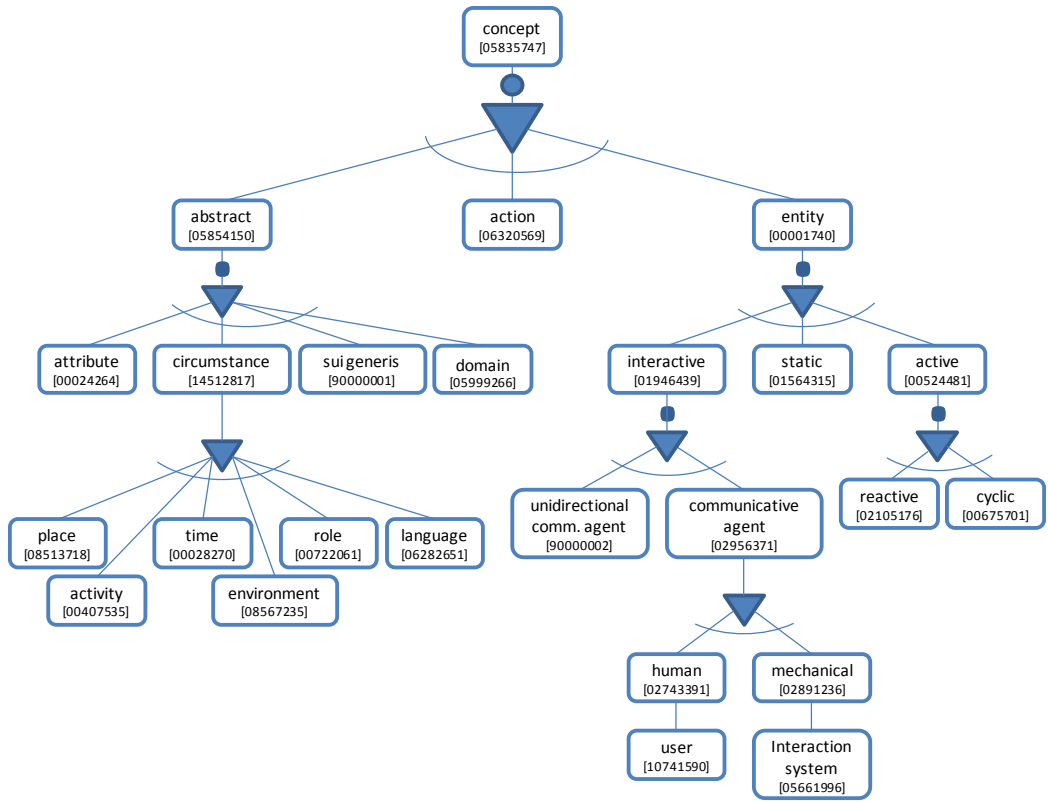

**Figure 3: Essential dimension taxonomy**

The *compositional dimension* represents the part-whole relationship between concepts. In this way, any concept can have relationships with a collection of concepts that are part of it. Figure 4 shows some of the concepts that are part of a computer, for example "hard disk", "RAM" and "ALU".

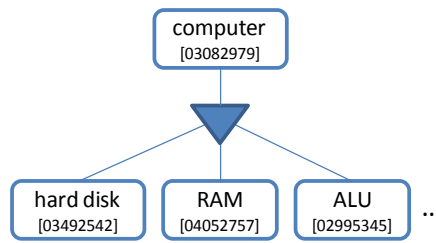

**Figure 4: Compositional dimension example**

The *restrictive dimension* shown in Figure 5 describes the compatibility between concepts related to some action and the rest. For example, the action "to compute" is related to the concepts "computer", "calculator" and "laptop", among others.

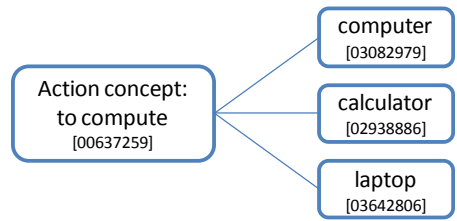

**Figure 5: Restrictive dimension example**

The *descriptive dimension* shown in Figure 6 is in charge of the relationships between three kinds of concepts: a generic concept (entity, abstract entity or action), an attribute likely to characterize that concept, and the domain (of values) on which that attribute is defined. Notice that there could be several available domains for a given attribute, and that a domain could be numeric (magnitudes regarding a unit) or enumerated (a concept which is *composed* of a set of named values which are also concepts). For example, an instance of the generic concept "hard disk" will have a value in the numeric domain of "information in bytes" for the attribute concept "storage capacity".

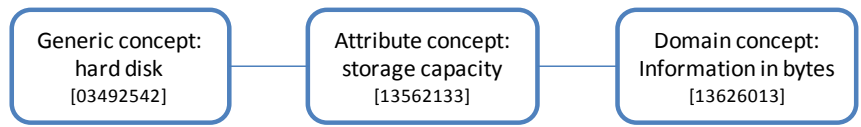

**Figure 6: Descriptive dimension example**

Finally, the *comparative dimension* is derived from previous dimensions and is responsible for calculating in real time the degree of similarity between ontological concepts. This paper, in fact, focuses precisely on that similarity calculation. Finally, for reasons of efficiency, most frequently requested similarities can be buffered, that is, stored when calculated, periodically updated and retrieved when necessary.

## 4. Proposal

This paper proposes and evaluates a similarity measure based on the combination of individual similarity measures according to each of the dimensions explained (see Section 3). This combination will be produced as training across numerous observations that will affect the weight with which each dimension contributes to the final decision. Training can be performed according to different criteria. On one hand, different human subjects support their judgments on different combinations of the dimensions. On the other hand, the nature of the concept determines the most relevant dimension for each comparison. For example, when comparing the concept 'scanner' with the concept 'printer', the sort dimension could be very influential, since both are types of computer peripherals; however restrictive dimension could not be as influential because they are related to different actions. The opposite may happen with the concepts 'teacher' and 'tutorial' because both are related to similar actions according to the restrictive dimension, such as 'teaching', while the sort dimension has little influence in this case.

The following step is to describe the similarity measure adapted to the described ontological dimensions except for the semiotic dimension. Yet not the only approach, similarity in the semiotic dimension, or similarity between terms is frequently described as the *edit distance* or Levenshtein distance (1966), that is, the number of changes necessary to turn one string into another string. The decision to leave this dimension apart is supported by preliminary studies in which this measure yields an average error rate above 50% and in some cases over 80%. Furthermore, for every concept in that study, the accuracy provided by this dimension was lower than that of some of the other dimensions (the semiotic dimension never produced the best prediction), being the only dimension which never ranked first when tested separately. For this reason, it is estimated as it cannot contribute positively to the results (at least, it cannot until it is properly adapted). Last but not least, during preliminary experimentation of the training including this dimension, it was observed that each weight tended to zero, and with the drawback of slowing down convergence of the weights of the rest of dimensions. However, as further work, some evolution of this similarity measure (supported by knowledge on this dimension) can be incorporated into the global measure of similarity.

## 4.1 Inference Mechanisms

This sub-section describes the method used to calculate the degree of similarity between two given concepts in an ontology. Since ontological knowledge here is structured into different dimensions, the similarity measure will also be based on these dimensions. Therefore, partial similarity calculations will be made for the sort, essential, compositional, restrictive and description dimensions described previously. The resulting overall similarity between the two concepts is obtained through the calculation of the weighted average of the five partial similarities

$$S = \frac{S_s w_1 + S_c w_2 + S_e w_3 + S_r w_4 + S_d w_5}{w_1 + w_2 + w_3 + w_4 + w_5}$$

where *Ss*, *Sc, Se, Sr* and *Sd* are the similarity measures according to the sort, compositional, essential, restrictive and description dimensions, respectively. The values *w1*, w*2*, *w3*, *w4* and *w5* represent the weights assigned to each dimension such that the resulting total similarity between the two concepts will be a value between 0 (completely different concepts) and 1 (the two concepts are the same).

The following sections describe in detail the procedures developed for the calculation of each of the partial similarities.

### 4.1.1 SIMILARITY ACCORDING TO SORT DIMENSION

The sort dimension represents the *is_a* relationship between concepts. This dimension has a polytree structure, allowing a concept to be a descendant of more than one concept. Similarity in this dimension is often calculated as proportional to the intersection of the list of predecessors of both compared concepts regarding the total size of these lists. To define this measure, a variation of the edge-counting technique – concretely, the conceptual similarity measure defined in the work of Wu and Palmer (1994) – has been employed. Given two concepts, $C_1$ and $C_2$, this measure can be defined as

$$Sim_s = \frac{2N_3}{N_1 + N_2}$$

where $N_1$ and $N_2$ are the number of ancestors of $C_1$ and $C_2$, while $N_3$ is the number of common ancestors of $C_1$ and $C_2$ (in the most advantageous tree if several are found in the polytree).

### 4.1.2 SIMILARITY ACCORDING TO COMPOSITIONAL DIMENSION

The compositional dimension represents the *part-whole* relationship between concepts. For this reason, the most appropriate way to calculate the similarity between two concepts based on this dimension is through the comparison of the parts (or ingredients) of these concepts. Furthermore, the calculation must also take into account the fact that a concept may consist of required and optional concepts. This detail is important when calculating similarity since a greater weight must be given to the required ingredients appearing in both concepts, while a lower weight is given to the optional ingredients. The resulting similarity of two concepts, $C_1$ and $C_2$, in terms of the compositional dimension is obtained by applying the formula:

$$Sim_c = \frac{N_1/M_2 + N_2/M_1 + 2N_3/(M_1 + M_2) + 2N_4/(M_3 + M_4)}{4}$$

where $N_1$ is the number of common components arising from the intersection of all components of concept $C_1$ with those components of concept $C_2$ of type *required*; $N_2$ is the number of common components arising from the intersection of all the components of $C_2$ with those *required* components of $C_1$; $N_3$ is the number of *required* components that both $C_1$ and $C_2$ have in common; and $N_4$ is the total number of common components (both *required* and *optional*) of the two concepts; $M_1$ and $M_2$ represent the number of *required* components of concepts $C_1$ and $C_2$, respectively. Finally, $M_3$ and $M_4$ indicate the total number of components that $C_1$ and $C_2$ have.

### 4.1.3 SIMILARITY ACCORDING TO ESSENTIAL DIMENSION

The essential dimension contains a set of abstract concepts which define generic types of concepts (such as action, entity, abstract, circumstance or attribute). This generic classification frequently influences human speakers when estimating similarity. Some other works on similarity calculation posed that concepts are only comparable if included in the same category of WordNet's taxonomy (RiTa.WordNet, 2008). Such approach endows a critical value to this dimension, while omitting the rest of the classification. What is proposed here is that this dimension can contribute to similarity estimation as any other (albeit with a certain weight that could be different than the rest), and that all the concepts observed in the design of the essential dimension may influence the similarity estimation.

The method for calculating similarity between two concepts $C_1$ and $C_2$ in the essential dimension is based on the intersection of their essential ancestors (ancestors within the subset of essential concepts). This is formalized as follows:

$$Sim_e = \frac{2Card(E_1 \cap E_2)}{Card(E_1) + Card(E_2)}$$

where $Card(E_1)$ and $Card(E_2)$ are, respectively, the total number of essential ancestors of concepts $C_1$ and $C_2$, while $Card(E_1 \cap E_2)$ indicates the number of common essential ancestors.

### 4.1.4 SIMILARITY ACCORDING TO RESTRICTIVE DIMENSION

The restrictive dimension is defined between a concept representing an action and another concept representing an entity. Similarity in this dimension is calculated in a different way depending on the type of concepts to be compared. For this reason, two different similarity measures exist for the dimension: comparing two actions and comparing two entities. Similarity between two concepts representing an entity will be based on the action concepts that both entities have in common. The formula used for the calculation of this similarity when comparing two entities, $C_1$ and $C_2$, is defined as

$$Sim_r = \frac{M_1/(N_1 + N_3) + M_2/(N_2 + N_4)}{2}$$

where $M_1$ and $M_2$ are the number of common actions that have a positive or negative restrictive relationship with the entities $C_1$ and $C_2$, respectively. The values $N_1$, $N_2$, $N_3$ and $N_4$ represent, respectively, the total number of actions having a positive relationship with the entity $C_1$, a negative relationship with $C_1$, a positive relationship with the entity $C_2$, and a negative relationship with $C_2$.

As regards the similarity between two concepts representing an action, this is calculated based on the set of concepts defined on these actions, being more similar the higher the number of restricted concepts in common. The formula to calculate the similarity between two action concepts ($C_1$, $C_2$) of a particular sign (positive or negative) is defined as

$$Sim_r = \frac{2N_3}{(N_1 + N_2)}$$

where $N_3$ is the number of common entities shared by the two actions, and $N_1$ and $N_2$ are the total number of entities having a restrictive relationship with $C_1$ and $C_2$, respectively.

### 4.1.5 SIMILARITY ACCORDING TO DESCRIPTIVE DIMENSION

The description dimension represents the relationship between a concept, an attribute and a value in a concrete domain. Similarity in this dimension is calculated differently depending on the type of concepts to be compared, that is, entities, attributes or domains. For pairs of concepts ($C_1$, $C_2$) representing an entity, the applicable formula is defined as

$$Sim_d = \frac{2N_1 + 2N_2 + N_3}{(M_1 + M_2)}$$

where $N_1$ is the number of common attributes without a default value assigned, $N_2$ is the number of common attributes whose value is the same for both entities and has not been assigned by default, and $N_3$ is the number of common attributes with the same value where one of them has been assigned by default. The terms $M_1$ and $M_2$ correspond to the total number of attributes related to the concepts $C_1$ and $C_2$, respectively.

If both concepts ($C_1$, $C_2$) are attributes, the formula to apply is defined as

$$Sim_d = \frac{2N_3}{(N_1 + N_2)}$$

where $N_3$ is the number of common values of both attributes, and $N_1$, $N_2$ is the total number of possible values which can have the attributes $C_1$ and $C_2$, respectively.

Finally, if the concepts to be compared ($C_1$, $C_2$) represent domains, the similarity according to this dimension is calculated based on the amount of common attributes (for which those domains apply) and the number of values shared by both domains.

$$Sim_d = \frac{{}^{2N_3}\!/_{(N_1 + N_2)} + {}^{2M_3}\!/_{(M_1 + M_2)}}{2}$$

where $N_3$ is the number of common attributes shared by the domains ($C_1$, $C_2$), and $N_1$, $N_2$ are the total number of attributes associated with them. Finally, $M_3$ is the number of common values defined in both domains, and $M_1$, $M_2$ are the total number of values of the two domains.

Finally, the concepts to be compared ($C_1$, $C_2$) may be values belonging to a domain, either enumerated or of a numeric type. For operating domains, it is necessary to define previously a correspondence between them. Numeric domains can be related through a function (typically, a lineal proportion). Relating an enumerated domain to a numeric domain can be achieved by assigning to each enumerated value a fuzzy label in the numeric domain. Finally, the correspondence between two enumerated domains always involves an intermediate numeric domain (with a correspondence defined to each of the two other domains). Once the values are comparable, the formula to measure their similarity is defined as follows:

$$Sim_d = 1 - \frac{|C_1 - C_2|}{|C_{inf} - C_{sup}|}$$

where $C_{inf}$ and $C_{sup}$ are, respectively, the lower limit and the upper limit within the range of values, and $C_1$ and $C_2$ are the correspondent numeric comparable values.

## 4.2 Preliminary Experimentation

Before testing the proposal, some preliminary experiments were performed to refine it and to obtain a first perspective on its validity. These experiments have been instructed on a set of similarity measures obtained from a total of 20 pairs of concepts evaluated by 17 human subjects. This dataset will be further described in Section 5.1.

Specifically, the individual influence of each dimension in similarity was tested thorough a set of experiments involving each of them separately. Since there is no combination of them, there is no need for training either. Figure 7 shows a box plot that represents the error measures produced individually for each dimension.

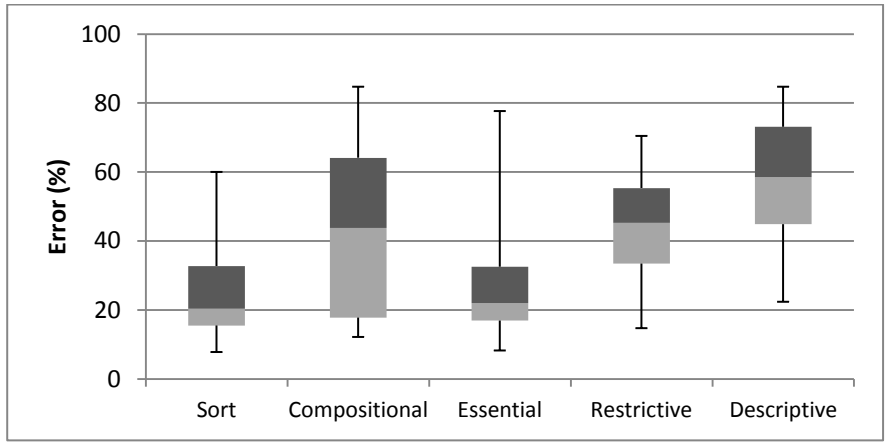

Figure 7: Performance of isolated dimensions of the Ontology

Figure 8 shows that in series of twenty pairs, every dimension produced better prediction than the others at least once. In fact, the essential dimension provided the best response in almost half of the cases, while the descriptive dimension was best in just one case.

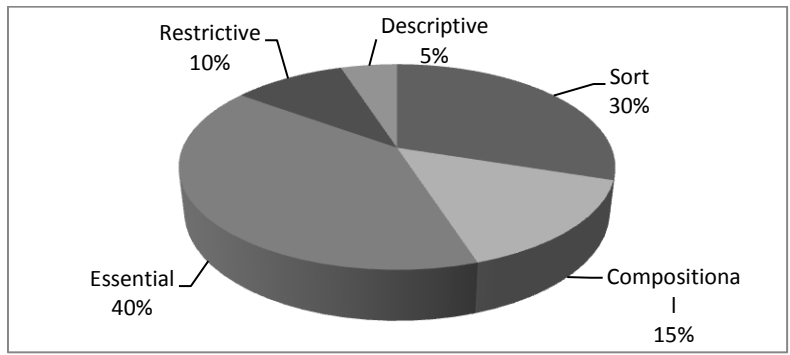

Figure 8: Cases in which each dimension is ranked first

This fact can lead to the conclusion that the essential design was appropriate, and that the descriptive dimension was weak. In further analysis it was found that the latter lacked sufficient knowledge, and it was improved in this line before evaluation (more knowledge was added). Despite this improvement, since the analysis and introduction of this knowledge is performed manually (in contrast to other dimensions, for which knowledge was obtained from WordNet), it could still be enhanced and this would improve the individual results of this dimension. Besides, this result is not definitive, since the weights may be different in other interaction domains, and the volume of the knowledge base is important too. But a useful consequence is that each one of the five ontological dimensions can contribute to the similarity function, supporting the hypothesis that an adequate combination of them may yield better results than any of these individual approaches.

### 4.3 Weights Training Methods

Assigning the proper weight to each dimension is crucial to achieving good results. Since a human test subject does not usually give the same relevance to the five dimensions of similarity, a basic training program regarding the weights associated with each dimension was developed. This program is based on the reinforcement learning technique (specifically a variant of the Q learning algorithm) and it has been implemented in order to determine, through several iterations, the appropriate value of the weights applied to each dimension (previously defined in Section 4.1) to minimize the error between the formula result and each human judgment. Therefore, the input to the training algorithm is the set of similarity judgments made by human test subjects. This algorithm follows the next steps:

    a) An initial step, where the five weights $w1$, w2, $w3$, $w4$ and $w5$ applied to each dimension (see formula in Section 4.1) are initialized to 1.

    b) For each iteration of the training algorithm, results for each dimension of similarity are calculated according to the formulas described in the Sections 4.1.1 to 4.1.5. Subsequently, the five new weights are calculated according to the next criteria:

        1. if $\forall i \; Sim_i < Y$,
           $\Delta w'_{\max (Sim_i)} = 1$

        2. if $\forall i \; Sim_i > Y$,
           $\Delta w'_{\min (Sim_i)} = -1$

        3. Failure to meet conditions 1) and 2),
           $\Delta w'_i = \alpha (Y - Sim_i) \, \Delta w_i \, Sim_i$

        where parameter $i$ is ranged from 1 to 5 (one for each dimension), $Sim_i$ represents each individual score and $Y$ represents a similarity value from 0 to 10 for one pair of concepts scored by one of the participant. $\Delta w'_i$ stands for the increase of the weight (for the dimension $i$) at the current iteration, while $\Delta w_i$ represents that increase at the previous iteration. The max$(Sim_i)$ and min$(Sim_i)$ represent the maximum and minimum similarity individual values, respectively. Finally, $\alpha$ stands for the learning rate.

    The training can be focused on different points of view, which will be tested and evaluated. Firstly, a pair-oriented training was implemented in order to individually adjust the weights for each of the 20 concept pairs, independently of the specific user. The weights are adjusted individually for each of the pairs of concepts, taking one user per iteration. In this way, after each iteration, a new array of refined weights is obtained and used for evaluating the similarity. The test consists of calculating the similarity (with that array of weights) and comparing it with the human assessment.

    Since the degree of significance assigned to each dimension may depend on the subjectivity of the testers, it was of particular interest to make an adjustment of the weights based on each user.

In this experiment, the training of the weights was performed once for each user and consisted of 20 iterations (one for each pair of concepts). For an iteration of this training algorithm, absolute error committed in relation to the corresponding pair was calculated. After running the training for the 17 users, the average of the absolute errors for each of the iterations was calculated.

The third method has been designed in order to address the shortcomings of the pair-oriented training. It should be indicated that storing an array of the weights for each possible pair of concepts in a medium sized ontology requires unusually extensive physical resources. Besides, a significant coverage of the thus defined knowledge would require far too much training. In short, it is not realistic to develop that method because of the high number of combinations of concepts. However, through preliminary experimentation it was checked that the weights applied to a pair were also likely to be applied to other combinations of each of those two concepts. Therefore, a new training method (*feature-oriented*) was proposed by slightly modifying the pair-oriented one. In the *feature-oriented* method, the array of weights is stored for each concept instead of for each pair of concepts (which solve both the problems of storage and the extent of training). Each time one concept is compared to any other, its array of weights will be reviewed and refined. The similarity calculation for a given pair is based on the aggregation of the arrays of both concepts.

Finally, it was observed that each method showed a different behavior depending on the pair of concepts compared: the method achieving the worst results on average was also the best for some specific pairs. Subsequently, a hybrid method was proposed and has been developed, combining the feature-oriented and user-oriented trainings, aiming to profit the advantages of each method. The training will be similar to that focused on the user, but for each iteration the array of weights will be refined to a different degree, taking into account the array stored for each particular concept. Therefore, if a particular dimension is usually relevant for a concept, adaptation to the user in that dimension will be strengthened.

## 5. Evaluation

Once the conceptual model of the ontology has been defined, and the weights training methods proposed, the next step in this study is to evaluate the proposal. The present section describes the experiments run for evaluating the proposal, from their design to the results obtained and discussion. The knowledge base is supported by the relational database management system Oracle 11g, and the logic of the ontology component (including the inference mechanisms) was implemented in Java. The knowledge bases were designed to satisfy specific purposes within a research project. The initial knowledge load was obtained from the large lexical database WordNet (Fellbaum, 1998) including all the existing concepts (synsets), terms and relationships (corresponding to sort and compositional dimensions). Since the proposed ontological model defines more relationships between concepts (essential, restrictive and descriptive), it is necessary to add more knowledge. The Cognos.Onto tool enables knowledge edition and management for this specific model. This tool belongs to a larger toolkit, Cognos (Calle et al., 2011) already used in several research projects. That toolkit seeks to ease the interaction corpus analysis, annotation, implementation and management, through diverse yet integrated tools aimed to each specific type of knowledge (pragmatic, NLP related, ontological etc.).

## 5.1 Experimental Design and Preparation

First of all, it is necessary to choose an Interaction Domain which will define the entire experiment. The concepts involved will be a subset of the whole knowledge base, restricted to that specific domain. The participants will be chosen in order to constitute a good coverage of the focused domain. Finally, additional knowledge will be fed by experts in that interaction domain not related to the projects where this research is framed (as the test subjects and any other participant in the experiments).

The methodology chosen to evaluate the proposed similarity measure is based on Miller's benchmark (Miller & Charles, 1991). Experiments have been designed to determine whether the result attained through the application of the similarity function on a pair of concepts is reliable or, in other words, if the result falls within an acceptable range when compared with the similarity judgments made by human test subjects.

To begin the experimental phase of the study, an initial loading of concepts must first be made in the proposed ontology. For this reason, WordNet's synsets (Princeton Univ., 2011) were taken as concepts, together with the corresponding semiotics, sort and compositional relationships. Knowledge domain experts have been responsible for populating the remaining dimensions of the ontological model (i.e., the essential, restrictive and descriptive) in a subset of 350 concepts, selected because of their relevance in the interaction domain.

The chosen domain is that labeled as "computer science teaching" interaction domain within the Spanish academic socio-cultural environment. This area of knowledge is familiar to the test subjects who have been selected as heterogeneous in this domain (different roles, ages, and genders). To perform the evaluation, a test was designed for which the test subject had to rate the similarity between pairs of concepts. The set of pairs had to meet a basic criterion: at least two pairs had to be included to explore each of the proposed dimensions, one with clear incidence in the dimension and another one without (or of little impact).

A total number of twenty-one test subjects were available, from which four outliers were left apart. They were discarded after checking their judgment because their responses were not uniform with the rest of the sample. The participant scores follow a normal distribution after removing the outliers. For that reason, the sample size was calculated through a test of statistical significance and the result was at least ten subjects to ensure a 99% confidence. Therefore, a sample size of seventeen participants is sufficient to ensure that the data is representative. The seventeen subjects were all experts in the interaction domain (technical education), specifically five technical students, seven researchers and five lecturers. Their ages ranged from 20 to 50 and were distributed as follows: seven subjects were in the 20-30 year-old range, six in the 30-40 year-old range and the remaining four were in the 40-50 year-old range. With regard to gender, slightly more than half of them were female (9) and the rest were male (8). The chosen interaction domain was the applied on the research project THUBAN (TIN2008-02711). Each participant was provided with a test containing a set of twenty pairs of concepts from this domain. Since the observations follow a normal distribution, it was determined that the minimum significant sample size would be sixteen with 99% confidence. Therefore, a set of twenty pairs of concepts provides significant results. However, in a larger domain, the size of the dataset may be different to attain statistically significant results. In coherence with some other components of the system where this proposal was to be integrated, the similarity measures are ranged from zero (no similarity) to ten (absolutely identical, the same concept). In addition, for each of the pairs, the subjects were asked to justify their score, indicating the specific parameters of similarity that they took into account in making their decision. After obtaining the individual survey results, the average total of the human assessments for each pair of concepts was calculated. Table 1 shows the 20 pairs of concepts included in the test and to the right of each pair, the *range* (difference between maximum and minimum scores), the standard deviation and the average rating assigned by the users.

| Pair ID | Pair of concepts | Range | Standard deviation | Average similarity |
|---------|------------------|-------|--------------------|--------------------|
| 0 | Reading lamp – Personal computer | 6 | 1.76 | 2.71 |
| 1 | Laptop – Server computer | 6 | 1.62 | 6.47 |
| 2 | Teacher – Tutorial | 7 | 1.92 | 5.06 |
| 3 | Meeting room – Laboratory | 8 | 2.15 | 4.35 |
| 4 | Server computer– Microwave | 8 | 2.02 | 2.24 |
| 5 | Office – Laboratory | 9 | 2.25 | 5.76 |
| 6 | Screen – Blackboard | 7 | 1.83 | 6.12 |
| 7 | Stapler – Folder | 7 | 2.19 | 3.94 |
| 8 | Plug– Power strip | 4 | 1.21 | 8.29 |
| 9 | Office – Meeting room | 6 | 1.69 | 6.29 |
| 10 | Pencil – CD marker | 3 | 0.99 | 7.29 |
| 11 | Associate professor – Teaching Assistant | 5 | 1.34 | 8.06 |
| 12 | Associate professor – Bachelor | 8 | 2.53 | 5.18 |
| 13 | To write papers – To program | 7 | 2.15 | 4.53 |
| 14 | To give a lecture – To teach | 6 | 1.60 | 7.76 |
| 15 | Keyboard – Mouse | 5 | 1.41 | 7.35 |
| 16 | Fridge – Microwave | 7 | 1.77 | 5.35 |
| 17 | Hard disk drive – Pendrive | 3 | 0.94 | 8.47 |
| 18 | Scanner – Printer | 8 | 1.89 | 5.94 |
| 19 | Poster – Blackboard | 6 | 1.82 | 4.24 |

**Table 1: Pairs of concepts and average similarity**

All the methods are subject to the iteration order (either analyzed pair or human judge), which can alter the result of the training. In order to avoid this effect and to endow significance to the results, through preliminary experiments the minimum number of repetitions (with different order) was determined to reduce stochastic and gain significance (close to 275), and consequently it was decided to program 300 repetitions with a different order for each method. In the graphs and tables, error rates of pairs (identified by *pair_id*) are numbered from 0 to 19, while iterations are numbered from 1 to 20.

## 5.2 The Experiments

This section presents the results obtained after the execution of the experiments corresponding to the four weight adjustment algorithms described in Section 4.3. These experiments were performed on a subset of the ontological knowledge stored acquired from the computer science teaching domain. The first experiment performed was the pair-oriented training and, in order to evaluate the results of this training, the average of the absolute error was calculated (for each pair) between the similarity based on each human judgment and the result obtained by applying the similarity measure proposed according to the following formula:

$$\overline{error_{pairId}} = \frac{\sum_{i=1}^{n} error_{pairId}(it_i)}{n}$$

where $i$ corresponds to an index to iterate over each human judge for a specific pair of concepts and $n$ is the number of test subjects. Finally, $error_{pairId}$ represents the absolute error between the human judgment for that pair and the result obtained through the training algorithm in that iteration. Table 2 shows the absolute errors calculated in this experiment for each pair of concepts, as well as the average error which, at about 18.5% comes slightly closer to the scores provided by the human subjects.

| | Pair Id | | | | | | | | | | | | | | | | | | | | AVG |
|---|---|---|---|---|---|---|---|---|---|---|---|---|---|---|---|---|---|---|---|---|---|
| | 0 | 1 | 2 | 3 | 4 | 5 | 6 | 7 | 8 | 9 | 10 | 11 | 12 | 13 | 14 | 15 | 16 | 17 | 18 | 19 | |
| error (%) | 15.2 | 14.8 | 38.3 | 18.6 | 19.4 | 18.1 | 17.6 | 18.8 | 20.2 | 15.4 | 13.4 | 18.0 | 22.5 | 19.6 | 15.2 | 13.0 | 15.3 | 20.9 | 17.1 | 19.0 | 18.5 |

**Table 2: Pair-oriented training error rate**

It should be noted that in eleven cases, the error rate is less than the average, in eight cases the error rate is around the average, and one pair (#2) shows an excessive error rate that requires further analysis and discussion (see subsection 5.3). Figure 9 shows a comparison of the trend lines regarding the error rate accumulated by the pair-oriented training algorithm and the accumulated error by the similarity function without weights training.

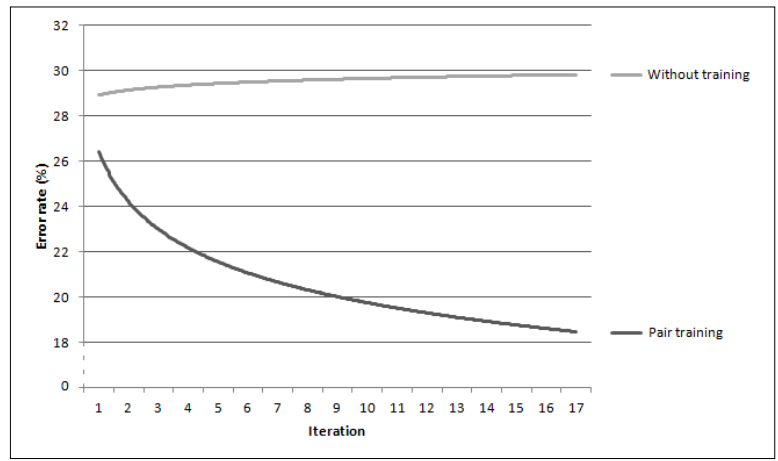

**Figure 9: Accumulated average error in pair-oriented training**

In second place, the absolute error obtained for each pair in the feature-oriented training is shown in Table 3. These results, compared with those obtained for the pair-oriented training, show slightly worse performance (with a mean error rate of 20,2%). However, it should be recalled that this method has other advantages (realistic storage and training extent).

| | Pair Id | | | | | | | | | | | | | | | | | | | | AVG |
|---|---|---|---|---|---|---|---|---|---|---|---|---|---|---|---|---|---|---|---|---|---|
| | 0 | 1 | 2 | 3 | 4 | 5 | 6 | 7 | 8 | 9 | 10 | 11 | 12 | 13 | 14 | 15 | 16 | 17 | 18 | 19 | |
| error (%) | 15.0 | 14.9 | 38.2 | 18.4 | 30.3 | 22.7 | 17.5 | 18.5 | 20.1 | 21.6 | 13.4 | 24.9 | 21.2 | 19.2 | 15.2 | 13.0 | 14.4 | 20.6 | 16.8 | 25.4 | 20.2 |

**Table 3: Feature-oriented training error rate**

The third experiment executed was the user-oriented training. In order to evaluate the results of this experiment, the average of the absolute error was calculated (for each human judge) between the similarity based on each human judgment for the 20 pairs of concepts and the result obtained applying the similarity measure proposed. In this way, the error average has been calculated as follows:

$$\overline{error_{userId}} = \frac{\sum_{i=1}^{n} error_{userId}(it_i)}{n}$$

where $i$ corresponds to an index to iterate over each pair of concepts for a specific user, $n$ is the number of pairs of concepts and $error_{pairId}$ represents the absolute error between the human judgment for that pair and the result of the training algorithm in that iteration.

In this case, the average error rate achieved is 23.9%, even worse than that for the feature-oriented training. The absolute error rate obtained for each iteration is shown in Table 4.

| | Pair Id | | | | | | | | | | | | | | | | | | | | AVG |
|---|---|---|---|---|---|---|---|---|---|---|---|---|---|---|---|---|---|---|---|---|---|---|
| | 0 | 1 | 2 | 3 | 4 | 5 | 6 | 7 | 8 | 9 | 10 | 11 | 12 | 13 | 14 | 15 | 16 | 17 | 18 | 19 | |
| error (%) | 18.6 | 14.1 | 40.7 | 17.9 | 30.8 | 16.8 | 22.9 | 17.5 | 37.1 | 17.1 | 34.6 | 35.8 | 24.3 | 21.5 | 31.2 | 13.6 | 13.9 | 27.4 | 22.3 | 20.8 | 23.9 |

**Table 4: User-oriented training error rate**

Figure 10 shows a comparison of the trend lines correspondent to the error rate accumulated by the user-oriented training algorithm and the accumulated error without any weight training. As can be observed, the user-oriented training trend line follows a downward curve and after 20 iterations reaches an error rate of 23.9%. Comparing both trend lines, it can be concluded that this training decreases the accumulated error and adapts the calculated similarities to the subject's judgments, yet it would be desirable to improve that adaptation (since it is still far from feature-oriented training).

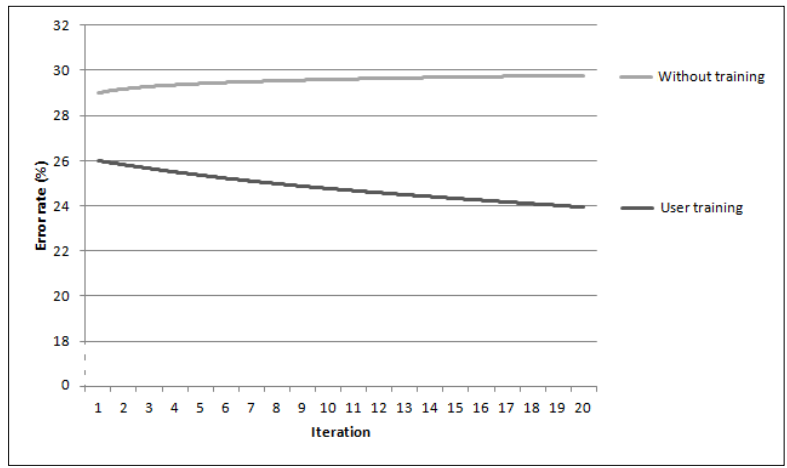

**Figure 10: Accumulated average error in user-oriented training**

As observed, both the user-oriented and the feature-oriented training methods are able to improve the similarities calculation, becoming noteworthy approaches. Consequently, it has been found of interest to explore a method which combines both of them. This new hybrid method departs from the user-oriented approach, and takes into account the weights vector obtained from the feature-oriented training described in section 4.3. As shown in Table 5, the user error rate has been successfully reduced to 21.2% with respect to the user-oriented training. However, this method degrades the performance achieved by the feature alone method.

| | Pair Id | | | | | | | | | | | | | | | | | | | | |
|---|---|---|---|---|---|---|---|---|---|---|---|---|---|---|---|---|---|---|---|---|---|
| | 0 | 1 | 2 | 3 | 4 | 5 | 6 | 7 | 8 | 9 | 10 | 11 | 12 | 13 | 14 | 15 | 16 | 17 | 18 | 19 | AVG |
| error (%) | 16.0 | 14.3 | 39.0 | 16.9 | 29.4 | 17.3 | 22.2 | 17.4 | 25.1 | 18.5 | 21.5 | 29.6 | 22.0 | 19.9 | 22.8 | 13.2 | 13.1 | 23.8 | 19.2 | 22.8 | 21.2 |

Table 5: User-feature hybrid training error rate

### 5.3 Discussion of Results Obtained

Among the results, concept pair 2 (*teacher-tutorial*) scored an error rate above 38% and the average similarity assigned by users (see Table 1) was 5.06. This latter value is significantly high considering the fact that the first concept refers to a person and the second is a static entity. Reviewing participant responses to this question, however, it can be understood that test subjects gave a higher score to the sole feature the concepts have in common, the activity of teaching. Analyzing the results of this outlier, it appears that the algorithm has a tendency to gradually increase the weight of the restrictive dimension, but longer training will be necessary to adapt the weight vector so that the only relevant dimension is the restrictive one. Using a training algorithm with faster convergence would ensure a good result in this pair, but could adversely affect the other results. However, convergence is guaranteed with a larger number of users.

Figure 11 shows the comparison of the absolute error obtained in the four experiments performed in this work (pair-oriented, user-oriented, feature-oriented and hybrid trainings) for each pair, and also the average results of each method. The first experiment performed, the pair-oriented training, achieves the best average error rate, about 18.5%, although in the pair mentioned above the error exceeded 38%. However, this experiment has a major limitation: a trained weight vector for each pair of concepts possible cannot be stored due to the large number of combinations of existing concepts in the ontology. This shortcoming was mitigated with the development of the feature-oriented training, achieving an error rate about 20.2%, a figure which is slightly worse than that of the pair-oriented training error. Nevertheless, this result does not fully reflect the impact of this training because not all test pairs include concepts that appear more than once in the experiment. If the calculation of the average error is restricted to those pairs which have concepts repeated in more than one pair, then the error amounts to 22.8%. In any case, this experiment has an important advantage since its implementation is more realistic and can be applied to large ontologies.

The user-oriented training was aimed at adapting the weights to each subject in order to confirm the assumption that not every test subject assigns the same value to all dimensions. Although the error rate achieved (23.9%) was not as satisfactory as either the pair or the feature-oriented trainings, the figure included in the sub-section 5.2 for the training shows a decreasing trend line which, when compared with the trend line without training, allows for the conclusion that the user-oriented experiment is able to adapt to each individual judgment. For this reason, an improvement was attempted with the user-training result through its combination with the feature-oriented experiment.

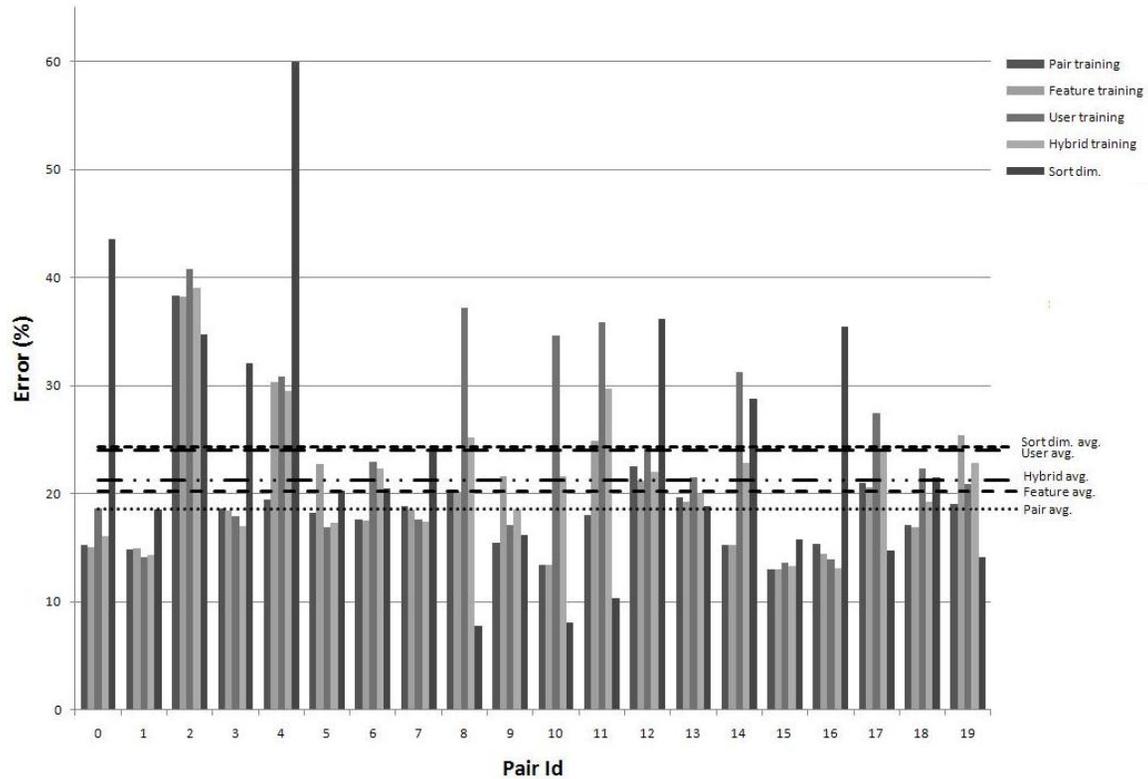

Figure 11: Comparison of the experiment results

The hybrid training detailed in Section 5.2 achieved 21.2% in the error rate, which reduces that of the user-oriented training, and balances the performance of the user-oriented method (reduces standard deviation). Taking into account that feature-oriented training method depends on the experience and that for some features the knowledge base might lack of this experience, the response obtained could not be satisfactory in some cases. In fact, when calculating the error produced by the feature-oriented method over the dataset (not restricted to repeated pairs) the result amounted to 22.8%. In sum, the feature-oriented method provides better results but only if enough knowledge is available. The last results presented in Figure 11 concern an experiment observing only the sort dimension (which is a frequent method for calculating similarities). Its average error rate is 24.1%, which is higher than that for any of the four methods discussed. In addition, it can be observed that the error rate of this experiment is, in several cases, far from the average error. Figure 12 shows a boxplot comparing the performance of the four training methods proposed and the sort dimension formula.

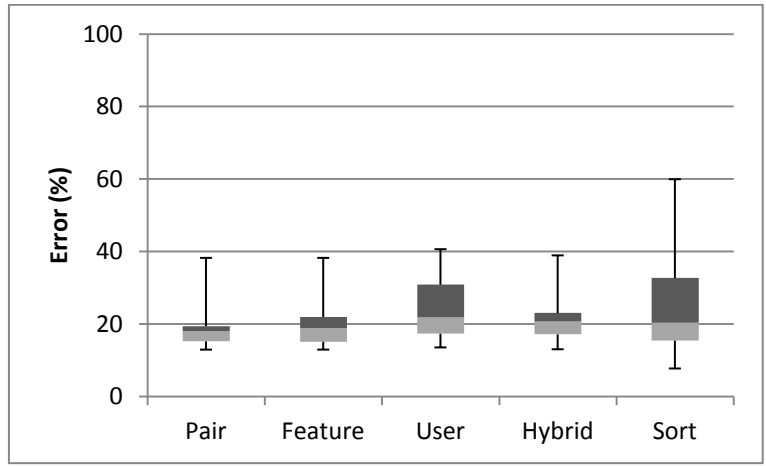

**Figure 12: Performance of each training method**

As can be seen, regarding the error in the predictions, the sort dimension obtains higher maximum (although also lower minimum), higher median (except for user training) and higher deviation than the rest. From this graph, it can be concluded that the error rate achieved by the sort dimension method (used in previous studies of similarity) is greater than the error rate achieved by the feature training method. In order to check that statistically, a null hypothesis was formulated (the average error is the same in both methods) and also an alternative hypothesis, (the average error of the feature-oriented method is lower than the sort dimension method error). The measure of discrepancy has been calculated for the sample of twenty measures of error (one per pair) and the result (-1.78) was found to be outside the acceptance range (-1.64, +∞), therefore the null hypothesis is rejected and the alternative is accepted with a significance level of 0.05. Consequently, it is considered true that the error shown by the feature-oriented method is lower than the error produced by the sort dimension method.

Finally, Figure 13 shows the average final weights of the four experiments. It shows the relevance taken through the experiments by each dimension, yet it cannot be extrapolated to other interaction domains. While dependent on the set of pairs chosen for the experiment, these results show that how all five dimensions are taken into account, with diverse weights.

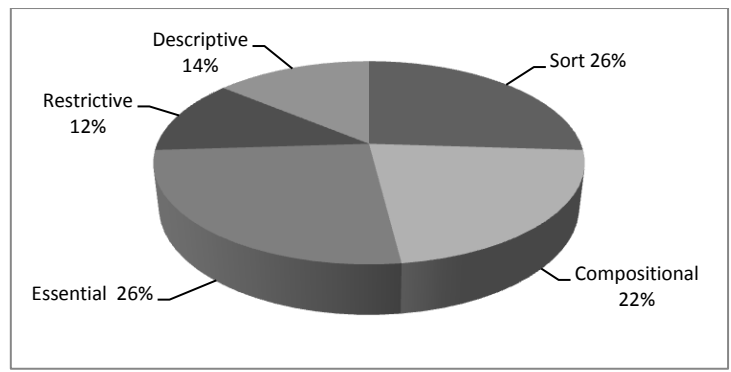

**Figure 13: Average weights of the ontological dimensions**

## 6. Conclusions and Perspective for Future Research

This paper defines a similarity measure for a multi-dimensional knowledge model of the ontology type, specifically an ontology aimed at supporting Human-Like Interaction. The proposed measure is based on five dimensions of ontological knowledge: sort, compositional, essential, restrictive and descriptive. The five of them are weighted and aggregated in order to obtain a global similarity measure. The equations applied for each dimension are general and can be used with other ontologies that observe any of these dimensions, yet observing all of them and aggregating their similarity result is here proposed for enhanced accuracy.

This solution presents another challenge, in the form of those weights calculation. In fact, when a person decides the similarity between concepts he unwittingly makes some dimensions prevail over the others. The criteria may be diverse, and this work has focused on studying the dependence of these weights on the nature of concepts, either in pairs (pair training method) or individually (feature training method), both described in Section 4.3. But this work also explores the influence of the past behavior of users who perform the concept pair evaluations (and ultimately, the user who owns a device or usually interacts with it). Following this line, a user-dependent training is proposed, and finally a hybrid one (merging feature and user benefits) is included too. All of them have been evaluated and compared in order to ascertain which one performs better, obtaining the best results for the pair-oriented training.

In order to evaluate the performance of the proposed similarity measure, its results were recorded and compared with those taken from human test subjects. This evaluation technique has been applied in several studies about similarity measures and is considered the gold standard. In the experimental phase, four training algorithms were developed according to different perspectives. Thus, this phase included a pair-oriented, a feature-oriented, a user-oriented and a hybrid experiment. In every case, the error rate was calculated with respect to the human subject assessments. The best results corresponded to the pair-oriented method which achieved an error rate of 18.5%. Since the implementation of this experiment is not realistic with large ontologies, a feature-oriented experiment was required despite slightly worsening the results from the previous experiment, concretely, producing an error rate of 20.2%. However, the feature-oriented experiment has the big advantage of being able to be applied easily to large ontologies.

Moreover, the user-oriented training aimed to adapt the weights to each subject in order to confirm the assumption that not every test subject assigns the same value to all dimensions. While this experiment had the highest error rate of all the algorithms (23.9%), as has been demonstrated, the error rate follows a decreasing trend line while, if training is not done, the error rate follows an asymptotic tendency. In addition to this, this experiment shows slightly better results than taking into account only the sort dimension (which has an average error rate of 24.1% and maximum 60.4%). For this reason, it can be concluded that the user-oriented experiment is able to adapt to each individual judgment (although this adaptation is very slow). Finally, the hybrid experiment combines the feature-oriented and the user-oriented training and, with an error rate of 21.2%, nevertheless manages to reduce the error of the user-oriented training, as well as balancing the error in the atypical cases common to the rest of the experiments.

Since the hybrid experiment manages to balance the results of the other experiments, currently, an improved hybrid algorithm is being developed. In this algorithm the calculation of the weights of each iteration will be affected depending on the error produced in the feature experiment for the pair of concepts corresponding to that iteration.

The performance of the training methods proposed is closely related to the available extent of knowledge. For this reason, authors are also currently working on mechanisms for increasing the quality and completeness of the ontological knowledge. The manual acquisition of new knowledge by an expert requires a great deal of resources and it would be desirable to develop an advanced mechanism to learn new concepts and relations. The challenge is to attain that knowledge acquisition through human-like interaction with human subjects. Therefore, through the lifetime of the system, the knowledge bases would be enriched by interacting with the users.

Finally, refinement of similarities formulation is also an interesting line of work, especially in the semiotic dimension for reintroducing its influence in the global similarity calculation.

## Acknowledgments

The development of this approach and its construction as part of the LaBDA-Interactor Human-Like Interaction System, part of the research projects SemAnts (TSI-020110-2009-419) and THUBAN (TIN2008-02711) and CADOOH (TSI-020302-2011-21), is supported by the Spanish Ministry of Industry, Tourism and Commerce and the Spanish Ministry of Education, respectively. Besides, the knowledge bases were populated using the COGNOS toolkit developed through the research project MA2VICMR (S2009/TIC-1542) supported by the Regional Government of Madrid.